%% file: main.tex
\documentclass[journal]{IEEEtran}

\ifCLASSINFOpdf
\else
   \usepackage[dvips]{graphicx}
\fi
\usepackage{url}

\usepackage{balance}
\usepackage{url}
\usepackage{graphicx}
\usepackage{booktabs, siunitx, tabularx}
\usepackage{cite}
\newcolumntype{C}{>{\centering\arraybackslash}X}

\usepackage{microtype}

\usepackage{amssymb}
\usepackage{amsmath}
\renewcommand{\vec}[1]{\boldsymbol{{#1}}}

\usepackage[prependcaption,textsize=scriptsize]{todonotes}
\definecolor{mycolor}{HTML}{FF6600}

\begin{document}

\title{Unsupervised Feature Learning for Speech using Correspondence and Siamese Networks}

\author{Petri-Johan Last, Herman A.\ Engelbrecht, and Herman Kamper}

\markboth{\copyright~2020 IEEE, Accepted to IEEE Signal Processing Letters}
{Last, Engelbrecht, Kamper}

\IEEEpeerreviewmaketitle

\maketitle

\input{abstract}

\input{introduction}
\input{herman_models}

\input{experiments}
\input{conclusion}

\newpage
\balance
\bibliographystyle{IEEEtran}
\bibliography{references}

\end{document}

%% file: abstract.tex
\begin{abstract}
In zero-resource settings where transcribed speech audio is unavailable, unsupervised feature learning is essential for downstream speech processing tasks. Here we compare two recent methods for frame-level acoustic feature learning. For both methods, unsupervised term discovery is used to find pairs of word examples of the same unknown type. Dynamic programming is then used to align the feature frames between each word pair, serving as weak top-down supervision for the two models. For the \textit{correspondence autoencoder} (CAE), matching frames are presented as input-output pairs. The \textit{Triamese network} uses a contrastive loss to reduce the distance between frames of the same predicted word type while increasing the distance between negative examples. For the first time, these feature extractors are compared on the same discrimination tasks using the same weak supervision pairs. We find that, on the two datasets considered here, the CAE outperforms the Triamese network. However, we show that a new hybrid \textit{correspondence-Triamese} approach (CTriamese), consistently outperforms both the CAE and Triamese models in terms of average precision and ABX error rates on both English and Xitsonga evaluation data.
\end{abstract}


%% file: introduction.tex
\section{Introduction}

\IEEEPARstart{A}{utomatic} speech recognition systems are usually developed using
large 
transcribed speech datasets.
However, for many languages it is difficult or impossible to collect the annotated resources required for training 
supervised speech recognition models~\cite{besacier+etal_speechcom14}.
In 
so-called \textit{zero-resource} environments, where unlabelled speech is the only available resource, unsupervised acoustic feature learning is essential for 
downstream 
tasks 
such as speech retrieval or 
indexing~\cite{zhang+glass_asru09, acoustic_indxing, Settle2017, wang+etal_icassp18}.
Features should ideally disregard irrelevant information (such as speaker and gender), while capturing linguistically meaningful contrasts (such as  phone or word categories).
Several different unsupervised frame-level acoustic feature learning methods have been developed over the last few years~\cite{jansen+dupoux,versteegh+thiolliere,versteegh+etal_sltu16,dunbar2017zero,heck+sakti,heck+etal_ieice18,dunbar+etal_interspeech19}, with neural networks being used in a number of studies~\cite{badino+etal_icassp14,badino+etal_interspeech15,renshaw+etal_interspeech15,tsuchiya,eloff+etal_interspeech19}.

One type of neural approach that has received particular attention is \textit{Siamese networks}~\cite{synnaeve+etal_slt14,thiolliere+etal_interspeech15,zeghidour+etal_icassp16,triamese,riad2018sampling}.
A Siamese network consists of
two identical sub-networks with tied weights taking in a pair of inputs~\cite{siamese}.
The standard Siamese network is trained so that the distance between the outputs of the sub-networks are minimised when the inputs are of the same type, and maximised when the inputs are of different types.
To train a Siamese network in an unsupervised fashion on unannotated speech, paired input is required.
Building on an idea from~\cite{jansen+etal_icassp13b}, Thiolli{\`e}re et al.~\cite{thiolliere+etal_interspeech15} used an unsupervised term discovery (UTD) system~\cite{park+glass_taslp08,jansen+vandurme_asru11} to automatically discover pairs of word-like segments predicted to be of the same type; these segments were aligned using dynamic time warping (DTW), giving matching frame pairs which were presented to the Siamese network as positive pairs.
Features from the resulting unsupervised Siamese network performed well in a minimal-pair triphone discrimination task~\cite{schatz+etal_interspeech13,versteegh+thiolliere}.

Siamese networks have subsequently been applied and further developed in a number of studies~\cite{zeghidour+etal_icassp16,triamese,riad2018sampling}.
Notably,~\cite{triamese}
introduced the \textit{Triamese network}, which takes triplets instead of paired 
input, with a triplet consisting of two positive items 
and one negative item. 
A contrastive loss 
is 
used
to minimise the distance between the two positive items such that it is smaller (by some margin) than the distance between a positive and 
negative item.
In discrimination tasks, it is often not the absolute distance between items that is of interest, but rather the relative distance---in contrast to the standard Siamese network, the Triamese network optimises relative distances.
Apart from showing that the Triamese approach is superior, \cite{triamese} also attempted to explicitly disentangle speaker and phonetic information within a single network (but this had limited effect).

Another type of neural approach that has seen success in learning acoustic features is the
\textit{correspondence autoencoder} (CAE)~\cite{cae}.
Instead of trying to reconstruct its input, as is done in a standard autoencoder~\cite{Goodfellow-et-al-2016}, the CAE tries to reconstruct another instance of the same type as its input.
To obtain input-output pairs from unlabelled speech,~\cite{cae} followed the same approach as used in the Siamese studies~\cite{thiolliere+etal_interspeech15}: a UTD system with subsequent DTW is used to find matching frame pairs to serve as weak supervision. 
By using the terms produced by the UTD system to train the CAE, the approach 
as a whole is unsupervised.
The CAE was also applied and extended in a number of subsequent studies~\cite{renshaw+etal_interspeech15,renshaw_masters16,code,hermann+goldwater_interspeech18,menon+etal_interspeech19,kamper_icassp19}.

Although the one uses a reconstruction-like loss while the other explicitly minimises or maximises a distance-based loss, both the CAE and Triamese networks rely on the same weak supervision from another unsupervised system.
Despite their commonalities, the performance of these networks has never been compared when trained on exactly the same paired data using the same evaluation metric.
The results in~\cite{renshaw+etal_interspeech15} and \cite{thiolliere+etal_interspeech15} are somewhat comparable (here the Siamese model seems to outperform an early version of the CAE), but the supervision pairs in the two studies are not the same and the feature dimensionalities are very different, so it is not possible to draw definitive conclusions.
In this paper we compare the performance of the two approaches in a controlled experimental setup: 
we train each network on the same discovered terms and evaluate the 
networks on the same tasks.
In this setting, we find
that the CAE outperforms the Triamese network.
More importantly,
we 
show that the two approaches 
are complementary and that they 
can be combined to form a \textit{correspondence-Triamese} (CTriamese) network.
This method 
consistently outperforms both the CAE and Triamese models in an evaluation on English and Xitsonga~data. 

%% file: herman_models.tex
\section{Unsupervised Speech Feature Learning}

We consider three approaches to unsupervised feature learning of frame-level acoustic features.
The first two (the CAE and Triamese models) were developed in previous work, while the last is a new hybrid approach.
All three approaches use weak top-down constraints in the form of discovered word pairs.
Concretely, we apply an unsupervised term discovery (UTD)
system~\cite{jansen+vandurme_asru11} to an unlabelled speech collection, resulting in pairs of word segments predicted to be of the same unknown type.
For each discovered word pair, dynamic time warping (DTW) is used to align the acoustic frames between the two words, yielding a set of frame pairs $\{ (\vec{x}_a, \vec{x}_b) \}$, 
where each $\vec{x} \in \mathbb{R}^D$ is an acoustic feature vector (e.g.\ MFCCs).
Each model below uses these frame pairs differently as weak supervision.

\subsection{Correspondence autoencoder (CAE)}
\label{sec:cae}

\begin{figure}[!b]
    \centering
    \includegraphics[scale=0.75]{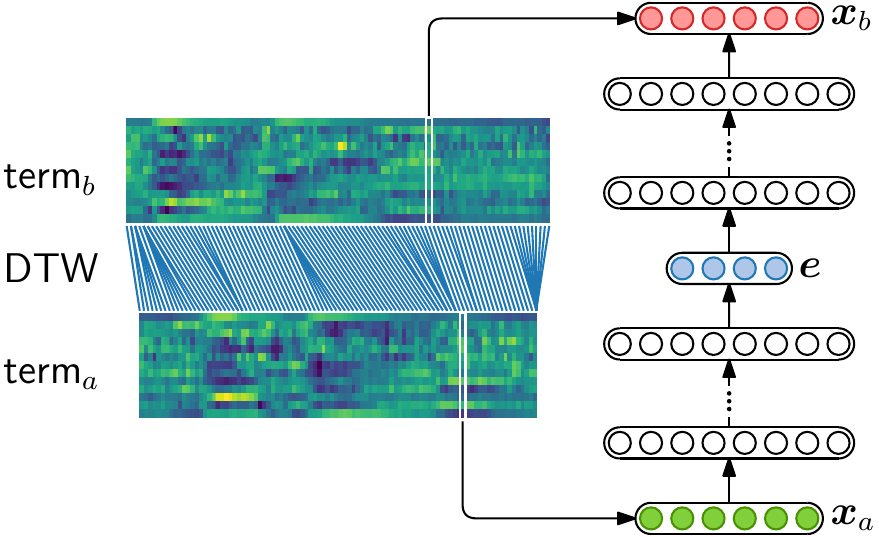}
    \caption{The correspondence autoencoder (CAE) is trained to reconstruct one acoustic frame $\vec{x}_b$ in a discovered word from another $\vec{x}_a$. The word pairs are discovered using unsupervised term discovery and aligned using DTW.}
    \label{fig:cae}
\end{figure}

Instead of using the same frames as input and output as in a standard autoencoder, the \textit{correspondence autoencoder} (CAE) uses the aligned frames as input-output pairs~\cite{cae}.
Formally, given an input frame $\vec{x}_a$, the CAE is trained using the loss $\ell_\textrm{cae} (\vec{x}_a, \vec{x}_b) = ||\hat{\vec{x}} - \vec{x}_b ||^2$, where {$\hat{\vec{x}}$} denotes the output of the network {and $\vec{x}_b$ the frame aligned to $\vec{x}_a$}.
The overall approach
is illustrated in Figure~\ref{fig:cae}.
The CAE network consists of a number of encoder layers, a 39-dimensional bottleneck layer, and decoder layers.
After training, we use the network up to the bottleneck layer $\vec{e}$ as our final feature extractor.

The idea behind the CAE is that it will result in features that are
insensitive to factors not common to the discovered word pairs, such as
speaker, gender and channel, while remaining dependent on factors that are, such as the word identity.

\subsection{Triamese network}
\label{sec:triamese}

An alternative to the reconstruction-like loss of the CAE is to train a model which explicitly learns to minimise a distance between paired inputs.
A \textit{Siamese network}~\cite{siamese} refers to a pair of networks with tied parameters which is trained to minimise the distance between two instances of the same type while maximising the distance between instances of different types.
This idea was first 
used for speech representation learning in~\cite{synnaeve+etal_slt14}.
In~\cite{triamese}, Zeghidour et al.\ extended this to \textit{Triamese networks}.
Rather than minimising or maximising an absolute distance, Triamese networks optimise the relative distance between instances of the same type and a negative example.

\begin{figure}[!b]
    \centering
    \includegraphics[scale=0.75]{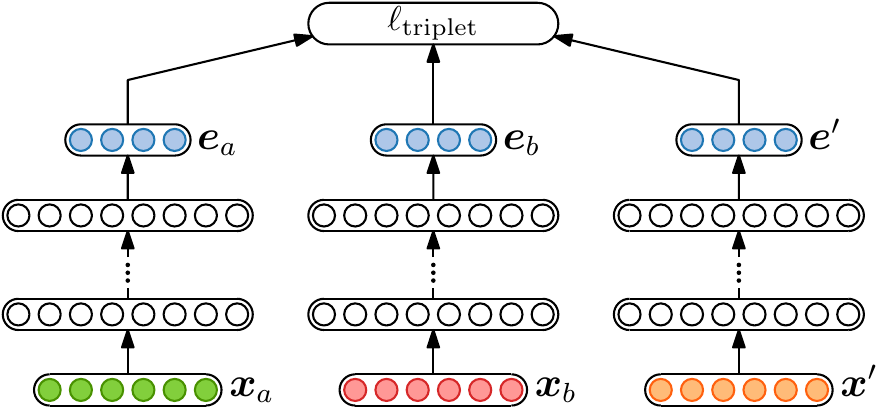}
    \caption{The Triamese network is trained so that the embeddings $\vec{e}_a$ and $\vec{e}_b$ of the same type are more similar by a margin $m$ than embeddings $\vec{e}_a$ and $\vec{e}'$ of different types.}
    \label{fig:triamese}
\end{figure}

Concretely, a Triamese network consists of three branches.
Parameters are shared between the branches, and each branch produces a feature embedding of its corresponding input.
Inputs consists of triples $(\vec{x}_a, \vec{x}_b, \vec{x}')$, where $\vec{x}_a$ and $\vec{x}_b$ are of the same type and $\vec{x}'$ is a negative example; the network produces corresponding embeddings $(\vec{e}_a, \vec{e}_b, \vec{e}')$.
The network is then trained using the following contrastive loss function~\cite{hermann+blunsom_iclr14}:
\begin{equation}
    \ell_\textrm{triplet}(\vec{x}_a, \vec{x}_b, \vec{x}') = \max \{ 0, m + d_{\cos} (\vec{e}_a, \vec{e}_b) - d_{\cos} (\vec{e}_a, \vec{e}') \}
    \label{eq:contrastive}
\end{equation}
where $d_{\cos}(\cdot, \cdot)$ denotes the cosine distance and $m$ denotes a margin parameter.
This loss is at a minimum when all embeddings $\vec{e}_a$ and $\vec{e}_b$ of the same type are more similar by a margin $m$ than embeddings $\vec{e}_a$ and $\vec{e}'$ of different types.
After training, one of the tied branches is used as a feature extractor.
The approach is illustrated in Figure~\ref{fig:triamese}.

The motivation for this approach is similar to that of the CAE in that we hope to obtain features that retain meaningful information while disregarding nuisance factors. However, while the CAE employs a soft reconstruction loss, the Triamese model optimises distances between representations discriminatively. The CAE is also only trained on positive pairs, while the Triamese model requires negative examples.
Here we sample negative examples randomly~\cite{triamese}, but~\cite{riad2018sampling} showed that more involved sampling schemes 
can give improvements. 

\subsection{Correspondence-Triamese network (CTriamese)}
\label{sec:ctriamese}

\begin{figure}[!t]
    \centering
    \includegraphics[scale=0.75]{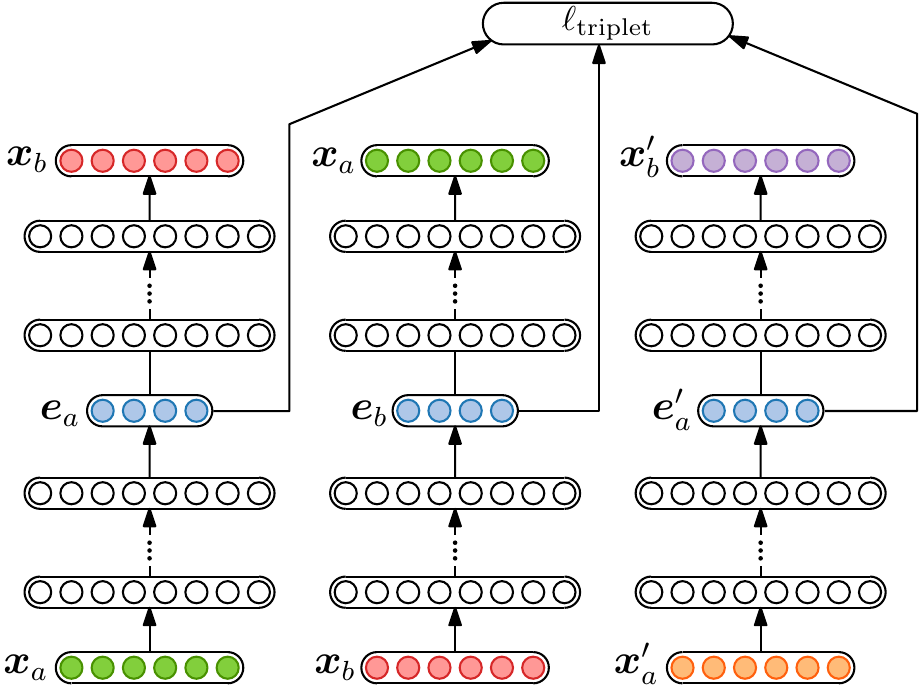}
    \caption{The correspondence-Triamese (CTriamese) network is trained using the combined loss of the three CAE branches and the contrastive triplet loss.}
    \label{fig:ctriamese2}
\end{figure}

Although the CAE and Triamese networks are both trained on discovered word pairs, they use this information very differently---the two approaches could therefore be complementary.
The \textit{correspondence-Triamese network} (CTriamese) applies the triplet loss function of the Triamese network to intermediate representations obtained from the CAE.
Concretely, each branch of the Triamese network above is replaced with a full CAE.
We then apply the triplet loss to the bottleneck layers from each CAE.
The full CTriamese network takes inputs $(\vec{x}_a, \vec{x}_b, \vec{x}'_a, \vec{x}'_b)$, where $(\vec{x}_a, \vec{x}_b)$ is a pair of the same predicted type, which is different from 
$(\vec{x}'_a, \vec{x}'_b)$.
The three CAE networks have the following respective input-output pairs: $(\vec{x}_a, \vec{x}_b)$, $(\vec{x}_b, \vec{x}_a)$ and $(\vec{x}'_a, \vec{x}'_b)$, as shown in Figure~\ref{fig:ctriamese2}.
Given the inputs, embeddings $(\vec{e}_a, \vec{e}_b, \vec{e}'_a)$ are produced and the network is trained with the following loss:
\begin{align*}
    \ell(\vec{x}_a, \vec{x}_b, \vec{x}'_a, \vec{x}'_b) &= \ell_\textrm{cae}(\vec{x}_a, \vec{x}_b) + \ell_\textrm{cae}(\vec{x}_b, \vec{x}_a) + \ell_\textrm{cae}(\vec{x}'_a, \vec{x}'_b) \\
    &\qquad + \ell_\textrm{triplet}(\vec{x}_a, \vec{x}_b, \vec{x}'_a)
\end{align*}

{Starting from a random initialisation, the parameters of the entire network is trained jointly to optimise this loss.}
For the CTriamese model, we also additionally include speaker information: 
after the bottleneck layer of each CAE, a speaker embedding is incorporated based on the {output} speaker identity for that CAE branch. These embeddings are learned jointly with the rest of the model and are concatenated to the first hidden layer after the bottleneck. This means that speaker information is not required at test time (when features are taken from the bottleneck layers).
In the experiments below, we incorporate speaker information in a similar way into the basic CAE.

%% file: experiments.tex
\section{Experimental Setup}

\subsection{Data}

We perform experiments on English and Xitsonga data.
For the English experiments, we use a subset of the conversational Buckeye corpus~\cite{buckeye}.
{As training data, around 12k terms from 12 speakers are discovered by applying the UTD system of~\cite{jansen+vandurme_asru11} to the complete set of unlabelled data.}
For validation and testing, around 3k and 4k true word segments are respectively used.
There is no speaker overlap between the training, validation or test data.
On 
English, 
early stopping on the validation set is performed to determine when to terminate 
training.

For the Xitsonga experiments, we use data from the NCHLT corpus~\cite{devries+etal_speechcom14}.
{Around 12k UTD terms from 24 speakers are discovered from the complete unlabelled training set, with 6.5k true words used for testing.
More details of the UTD system can be found in~\cite{versteegh+thiolliere}, where pair-wise matching precisions of 39\% and 69\% are reported on English and Xitsonga, respectively.}
The optimal hyperparameters determined on the English validation data are used directly on Xitsonga without any alteration, representing the realistic setting where we have a zero-resource language without any labelled validation data.
Since validation data is also not available for early stopping, the Xitsonga models are trained for the 
number of epochs that was found to be optimal on the English validation data.

We use
cepstral mean and variance normalised Mel-frequency cepstral coefficients (MFCCs) with first- and second-order derivatives to represent the speech audio. 
The discovered word pairs are DTW-aligned
to obtain frame-level pairs for training the different networks.
For the Triamese network, which also requires negative examples, 
we randomly choose a frame from a third discovered word that is spoken by the same speaker as the first word in the pair, but with a different predicted type. 
The CTriamese network requires a fourth item 
for the final CAE branch's output.
We randomly choose a fourth word that matches the predicted type 
of the third word in the triplet, and DTW-align it with the third word to form the 
input-output frame-pair for the third network branch.

\subsection{Neural network architectures}

\subsubsection{Correspondence autoencoder}
The CAE (\S\ref{sec:cae}) 
consists of two halves: an encoder and a decoder.
In our case, the encoder and decoder 
both consist of six 100-unit ReLU layers with a 39-dimensional middle bottleneck layer from which features are extracted.
The network is trained using the Adadelta optimiser \cite{zeiler2012adadelta} with a learning rate of 0.001, as was done in~\cite{cae}. 

\subsubsection{Triamese network}
A single branch in our Triamese network (\S\ref{sec:triamese}) consists of six densely connected 100-unit ReLU layers followed by a 39-unit ReLU 
embedding layer.
\cite{triamese} 
used four hidden layers of 1000 units each and an embedding layer 
of 100 units.
We use a different architecture to keep embeddings comparable (39 dimensional), but include development results using the architecture of~\cite{triamese}.
As in \cite{triamese}, the Triamese network is trained using stochastic gradient descent with a learning rate of 0.01 and a decay of $\textrm{10}^{-\textrm{6}}$ to optimise
the contrastive loss in~\eqref{eq:contrastive} with a margin of $m = 0.15$.

\subsubsection{Correspondence-Triamese network}
Each branch of the CTriamese network (\S\ref{sec:ctriamese}) mirrors the structure of a single CAE, consisting 
of six 100-unit ReLU encoding layers,
a 39-unit bottleneck embedding layer, and another
six 100-unit ReLU decoding layers.
For the contrastive loss, a margin of $m = 0.15$ was found to perform best on the English validation data.
For adding speaker information in the CAE and CTriamese models, we use a speaker-embedding dimensionality of 100.
The CTriamese network is trained using Adadelta \cite{zeiler2012adadelta} 
with a learning rate of 0.001.
{On our data, each model took less than an hour to train on a single GeForce GTX~1080 GPU.}

\subsection{Evaluation}

The multi-speaker word discrimination task developed in~\cite{evaluation}, called the \textit{same-different task}, quantifies how similar words of the same type are relative to other words of different types based on 
a given speech representation.
This task has been developed specifically for evaluating speech features without requiring downstream modelling assumptions.
By calculating the DTW distance between every pair in a set of isolated test words (represented using the features under evaluation) 
we classify two words as being the same
if the distance between them is less than some threshold. 
The threshold is then varied from zero to the maximum 
to produce a precision-recall curve.
The area under this curve is called the average precision (AP), which we use as our first evaluation metric.

We also consider ABX discrimination, a task inspired by human perceptual studies asking whether the distance between an anchor item $A$ and an another item $X$ of the same type is smaller than the distance between the anchor and an imposter $B$ according to the representation under evaluation~\cite{schatz+etal_interspeech13}.
We specifically use the cross-speaker triphone discrimination task~\cite{versteegh+thiolliere}, where triphone $A$ and $B$ are minimal pairs differing in the middle phone (e.g.\ `bag' and `bug'), with $A$ and $B$ from the same speaker and $X$ from a different speaker. ABX is reported as an error rate averaged over all minimal triphone pairs in the corpus.\footnote{We use the most recent version of the ABX evaluation code~\cite{schatz2016abx}. Because of changes, this means that results might not be directly comparable to previous studies, but performance across the models here can be compared directly.}

\section{Experimental results}

Table \ref{tbl:resultsutd} gives results on the English and Xitsonga validation and test sets for 
unsupervised models 
trained on UTD-discovered terms.
We first evaluated the AP of all the models on the English validation set, and then evaluated only the most representative models on the English and Xitsonga test data
(which is why test results are missing for three model variants).
Speaker conditioning is indicated with the `w/~speaker' label.
A larger AP indicates that the features allow for better word discrimination while a lower ABX error rate indicates better minimal-pair triphone discrimination.
We also include a baseline where the same-different and ABX tasks are performed directly using DTW over MFCCs.

Table \ref{tbl:resultsutd}  indicates that
the CAE outperforms the Triamese networks
on both languages.
While the CAE outperforms the MFCC baseline in all cases, the Triamese model with a 39-dimensional embedding layer fares worse than MFCCs on the same-different task (AP) on the English test data.
To show that this is due to the lower dimensionality used here, we include development results where a 100-dimensional embedding is used with the architecture of~\cite{triamese}.
This model achieves a 13\% relative improvement in AP over the MFCCs.
Moreover, despite the lower dimensionality, the 39-dimensional Triamese model still outperforms MFCCs in the ABX task on both languages (as reported in previous work).
Future work could consider architectural improvements and more advanced Triamese-pair sampling schemes~\cite{riad2018sampling},  but our specific goal here is a fair like-for-like comparison between the different models, which is why we use the 39-dimensional Triamese model for testing.

Considering the results of the CAE and CTriamese networks with 
and without speaker conditioning, we see that speaker
conditioning slightly worsens the performance of the CAE while the 
CTriamese benefits from speaker conditioning.
In \cite{triamese}, a method for incorporating speaker information in the Triamese model was also proposed, 
but this did not consistently improve discrimination performance.

The best performance across all sets and languages are achieved by the hybrid CTriamese network, outperforming the MFCC baseline and both the CAE and Triamese networks in terms of AP and ABX error rate. This shows that the CAE and Triamese network's methodologies are complementary.

\begin{table}
\caption{Average precision (AP) and ABX error rate (both in \%) 
for unsupervised neural network feature extractors trained on terms from an unsupervised term discovery (UTD) system.
}
\begin{tabularx}{\columnwidth}{l@{}cCCCC}
\toprule
&
\multicolumn{1}{c}{ English val.} &
\multicolumn{2}{c}{ English test} &
\multicolumn{2}{c}{ Xitsonga test} \\
\cmidrule(l){2-2} \cmidrule(l){3-4} \cmidrule(l){5-6}
{Model} &
AP & AP & ABX & AP & ABX \\
\midrule
MFCC & 36.8 & 35.9 & 21.4 & 28.1 & 19.0 \\
\addlinespace
CAE & 47.4 & 46.0 & 16.9 & 57.4 & 12.1 \\
CAE w/ speaker & 47.2  & - & - & - & - \\
\addlinespace
Triamese (39 dim.) & 38.2  & 34.8 & 20.7 & 45.9 & 17.6 \\
Triamese (100 dim.)~\cite{synnaeve+etal_slt14} & 41.5  & - & - & - & - \\
\addlinespace
CTriamese & 49.3  & - & - & - & - \\
CTriamese w/ speaker & \textbf{50.4} & \textbf{47.2} & \textbf{16.5} & \textbf{60.7} & \textbf{12.0} \\
\bottomrule
\end{tabularx}
\label{tbl:resultsutd}
\end{table}

\begin{table}
\caption{
Average precision (\%) on English data 
when training the models on ground truth words instead of UTD-discovered terms.}
\centering
\renewcommand{\arraystretch}{1.2}
\begin{tabularx}{0.7\columnwidth}{lCC}
\toprule
Model &
Validation & Test \\
\midrule
MFCC & 36.8 & 35.9 \\
CAE & 51.9 & 45.5 \\
Triamese (39 dim.) & 46.3 & 38.8 \\
CTriamese w/ speaker & \textbf{53.0} & \textbf{50.8} \\
\bottomrule
\end{tabularx}
\label{tbl:resultsgt}
\end{table}

In Table \ref{tbl:resultsgt} we also present the same-different performance of the models when trained on ground truth English words instead of discovered terms
in order to quantify
the performance loss induced by the use of UTD terms in the truly unsupervised setting (Table~\ref{tbl:resultsutd}).
While the trends are the same as in Table~\ref{tbl:resultsutd}, 
the Triamese network sees a 
larger relative increase in performance when trained on the ground truth words. 
This seems to indicate that the soft reconstructive loss of the CAE is less sensitive to the quality of the training data than the contrastive loss of the Triamese network.
The difference in performance between the CAE and CTriamese approaches are also smaller in Table~\ref{tbl:resultsgt} (when using ground truth words) than in Table~\ref{tbl:resultsutd} (the unsupervised case).

%% file: conclusion.tex
\section{Conclusion}
We have compared the performance of the correspondence autoencoder (CAE) and Triamese networks on word and minimal-pair triphone discrimination tasks.
We found that the CAE outperforms our implementation of the Triamese network.
More importantly, we showed that
the methodologies of the Triamese and CAE networks are 
complementary: we combine the two into a new CTriamese hybrid network, which outperforms both the individual 
approaches in intrinsic evaluations on English and Xitsonga data.
In future work we plan to use the CTriamese approach in 
downstream speech processing tasks, and to incorporate some of the more advanced Triamese-pair sampling strategies that have been proposed~\cite{riad2018sampling}.

\section*{Acknowledgements}

This work is based on research supported in part by the
National Research Foundation of South Africa (Grant Numbers: 120409). HK is supported by a Google Faculty Award.